\documentclass[letterpaper, 10 pt, conference]{ieeeconf}  

\IEEEoverridecommandlockouts                              

\usepackage{url}
\usepackage{graphicx}
\usepackage{booktabs}
\usepackage{amsmath}
\usepackage{amsfonts}
\usepackage[table]{xcolor}
\usepackage{longtable}
\usepackage{hyperref} 
\usepackage{balance}
\usepackage{comment}
\usepackage{array}
\newcolumntype{P}[1]{>{\centering\arraybackslash}p{#1}}
\title{\LARGE \bf Sim-to-Real Gentle Manipulation of Deformable and Fragile Objects with Stress-Guided Reinforcement Learning
}

\author{Kei Ikemura$^{1*}$, Yifei Dong$^{1*}$, David Blanco-Mulero$^{2}$, Alberta Longhini$^{1}$, Li Chen$^{1}$, Florian T. Pokorny$^{1}$
\thanks{The authors are with: $^{1}$the division of Robotics, Perception and Learning, KTH Royal Institute of Technology, Stockholm, Sweden;
$^{2}$Institut de Robòtica i Informàtica Industrial, CSIC-UPC, Barcelona, Spain.
Funded by the European Commission under the Horizon Europe Framework Programme project SoftEnable, grant number 101070600. Contact: {\tt\small \{ikemura, yifeid\}@kth.se}.}}

\begin{document}

\maketitle
\def\thefootnote{*}\footnotetext{Equal contributions.}

\thispagestyle{empty}
\pagestyle{empty}

\begin{abstract}
Robotic manipulation of deformable and fragile objects presents significant challenges, as excessive stress can lead to irreversible damage to the object.
While existing solutions rely on accurate object models or specialized sensors and grippers, this adds complexity and often lacks generalization.
To address this problem, we present a vision-based reinforcement learning approach that incorporates a stress-penalized reward to discourage damage to the object explicitly.
In addition, to bootstrap learning, we incorporate offline demonstrations as well as a designed curriculum progressing from rigid proxies to deformables.
We evaluate the proposed method in both simulated and real-world scenarios, showing that the policy learned in simulation can be transferred to the real world in a zero-shot manner, performing tasks such as picking up and pushing tofu.
Our results show that the learned policies exhibit a damage-aware, gentle manipulation behavior, demonstrating their effectiveness by decreasing the stress applied to fragile objects by $36.5\%$ while achieving the task goals, compared to vanilla RL policies. 
\end{abstract}


\section{INTRODUCTION}
Deformable and fragile objects manipulation (DFOM) requires gentle and reliable handling of the object, which is crucial in applications ranging from food processing and agriculture to assistive care and surgery~\cite{wang2022challenges,swann2025dexfruit,chua2021toward,masui2024vision,wang2025image}.
Manipulation of these objects poses major challenges, as objects can tear, crack, or bruise under excessive compression or tension~\cite{zhu2025deformable,blanco2024t}. The difficulty stems from two factors: (i) complex contact and object dynamics that are hard to model, making it difficult to assess the internal stress experienced by the object, and (ii) stringent safety constraints that require keeping the stress below damage thresholds~\cite{yin2021modeling,arriola2020modeling}.

The problem of DFOM can be cast as minimizing the internal stress applied during interaction while still achieving task goals. Model-based methods approach this problem by leveraging physical priors and analytical models to explicitly measure the stress levels~\cite{pan2020grasping, huang2023defgraspnets}. However, such approaches typically require accurate object models and precise parameter identification, making them difficult to deploy in the real world. This motivates the use of reinforcement learning (RL), which has demonstrated strong performance in visuomotor control tasks through sim-to-real transfer~\cite{tang2024deep, matas2018sim}. 
Despite this promise, applying RL to real-world DFOM remains largely unexplored, especially given the challenge of ensuring safe stress levels in fragile objects. 

\begin{figure}
    \centering
    \includegraphics[width=0.995\linewidth]{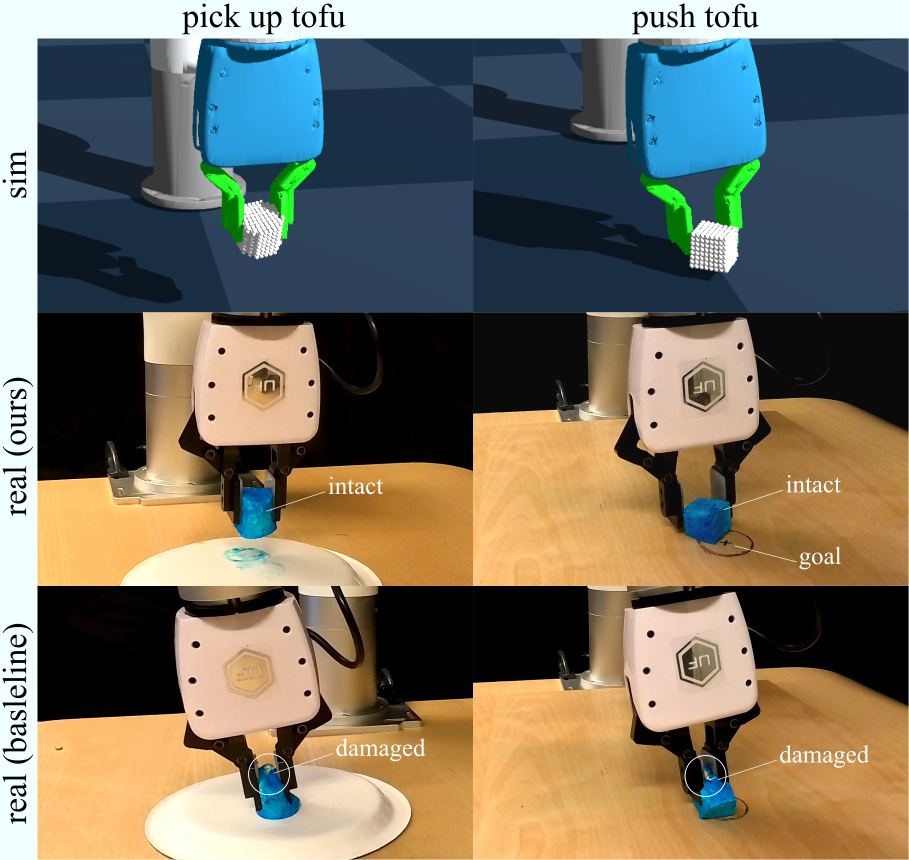}
    \caption{Overview of our approach to deformable and fragile object manipulation using only visual input. Our stress-guided RL policy, trained in simulation, transfers to the real world zero-shot. It enables tasks such as picking up and pushing tofu without causing damage, in contrast to a baseline. For visualization, the tofu is dyed blue in real-world experiments.}
    \vspace{-7pt}
    \label{fig:front}
\end{figure}


In this work, we address this gap by introducing a stress-guided RL framework specifically designed for DFOM tasks (Fig.~\ref{fig:front}), instantiated on a standard off-policy RL method~\cite{luo2024serl}. The policy is trained from point cloud observations obtained with an RGB-D camera, which capture object geometry and deformation in a form well suited for sim-to-real transfer, without requiring precise object models or additional sensors (e.g., tactile sensors).
To explicitly account for fragility, we design a stress-penalized reward computed in simulation, which encourages gentle manipulation behaviors by penalizing actions likely to cause damage.
While this reward captures safety constraints, it inherently conflicts with task success, making the vanilla RL converge slowly or fail to converge altogether.
To address this, we introduce two mechanisms that facilitate stable learning: (i) a curriculum that progresses from rigid proxies to deformable objects, and (ii) the use of offline demonstrations for policy bootstrapping.
We comprehensively evaluate our method in both simulated and real-world scenarios. The method demonstrates strong effectiveness in decreasing the stress applied to the fragile objects by $36.5\%$ compared to a vanilla RL policy, while succeeding in the manipulation task.

In summary, our main contributions are:
\begin{itemize}
    \item We propose the first visuomotor learning framework for sim-to-real manipulation of 3D deformable and fragile objects, which explicitly incorporates object fragility into the learning process.
    \item We introduce a stress-penalized reward that enables safe and gentle handling of deformable and fragile objects.
    \item We incorporate offline expert demonstrations and curriculum to learn from the gentle manipulation behavior of humans and bootstrap the learning process.
    \item We demonstrate zero-shot sim-to-real transfer in two challenging tasks involving picking and pushing tofu with minimal damage to the object.
\end{itemize}

\section{RELATED WORKS}
\subsection{Deformable and Fragile Object Manipulation}
Existing approaches to DFOM typically exploit specialized hardware, advanced sensing, or policy learning techniques.
Hardware solutions often involve soft or compliant end-effectors designed to reduce the risk of damage to the object~\cite{maruyama2013delicate,wang2021circular}.
However, such designs require substantial engineering effort and task-specific expertise.
Alternatively, methods leveraging tactile, force, or touch sensors enable precise control and safe interaction.
For example, Huang et al.~\cite{huang2019learning} train policies for gentle contact with fingertip touch sensors, trained in simulation and real world. Similarly, Yagawa et al.~\cite{yagawa2023learning} propose a framework to learn to anticipate fracture during grasping with tactile feedback, and Lee et al.~\cite{lee2025grasping} study deformable grasping using visuo-tactile simulation. Lee et al.'s approach has been validated in simulation, though real-world evaluation remains to be studied.
While promising, these methods using additional on-finger sensors often increase the system complexity and cost.  

To avoid the issues, an alternative approach is to utilize vision-based strategies to estimate interaction forces from vision.
Wang et al.~\cite{wang2025image} propose a framework that predicts contact forces using structured-light.
Jung et al.~\cite{jung2020vision} estimate suture tension from video and tool poses.
Similarly, Masui et al.~\cite{masui2024vision} predict manipulation forces from surgical images to detect unsafe over-force events.
In contrast, our work leverages stress information computed in simulation as a physical prior. With approximate knowledge of material properties, a stress-penalized policy can be learned in simulation and transferred to the real world through domain randomization.
This avoids reliance on specialized tactile sensors while exploiting the fidelity of modern soft-body simulators to achieve safe DFOM~\cite{zhou2024genesis}.  

\subsection{Sim-to-Real Reinforcement Learning for Manipulation}
Sim-to-real reinforcement learning has been widely explored in diverse domains, including in-hand manipulation~\cite{andrychowicz2020learning}, dexterous manipulation~\cite{lin2025sim}, tactile-based manipulation~\cite{lin2023bi}, and manipulator co-design~\cite{dong2024cagecoopt}. For deformable object manipulation, Matas et al.~\cite{matas2018sim} pioneered training RL agents entirely in simulation and transferring the learned policies to the real world using only visual input. Subsequent works often rely on reward shaping to guide learning. For instance, curiosity-based rewards have been applied to minimize impact forces and promote safe exploration~\cite{huang2019learning}, while sparse rewards have been used to achieve consistent grasping of deformable food items such as spaghetti~\cite{prem2025autonomous}. Although effective, such strategies demand extensive reward engineering. Other approaches integrate curriculum learning with RL to progress from simpler to more complex tasks, including cable insertion and robotic surgery~\cite{narvekar2020curriculum,scheikl2022sim,wu2025robotic}. 
Recent advances further combine RL with offline demonstrations and human-in-the-loop interventions, improving sample efficiency and training stability significantly~\cite{luo2024serl,luo2025precise}. These methods have proven effective in tasks such as deformable timing-belt assembly.  This line of sample-efficient RL methods is particularly appealing to sim-to-real transfer in DFOM, where the computational demands of soft-body simulation pose major challenges.
Inspired by these developments, we investigate sim-to-real reinforcement learning for 3D deformable and fragile object manipulation that explicitly accounts for object fragility - a previously unexplored research gap.
Our method couples stress-guided rewards with curriculum learning or demonstration bootstrapping. It aims to enable safe and gentle manipulation policies under real-world fragility constraints.




\section{Preliminaries}
In this work, we study the manipulation of deformable and fragile objects with two objectives: (i) successfully performing the task, such as picking up or pushing an object to a target, and (ii) minimizing applied stress to ensure object is not damaged. We begin by formally defining the problem (Section~\ref{sec:prob_form}), followed by an introduction of key concepts and notations related to object stress (Section~\ref{sec:method-prelim-stress-metrics}), which form the basis of our method.

\subsection{Problem Formulation}\label{sec:prob_form}
We consider the problem of learning a DFOM policy that achieves the aforementioned objectives.
That is, controlling a robot end-effector to perform a manipulation task while minimizing damage to the object.
Here, we assume partial observability of the state of the object. 
Thus, the stress applied to the object can only be measured in simulation.

Formally, we model DFOM as a partially observable Markov decision process (POMDP). The state space $\mathcal{S}$ encodes the joint position and velocities of the robot, as well as the particle positions and velocities of the object and its internal stress. The observation space $\mathcal{O}$ includes a low-dimensional encoding (via neural network feature extractors) of the object’s partial point cloud~$\mathbf{o} \in \mathbb{R}^{d}$, the centroid of the point cloud~$\mathbf{c} \in \mathbb{R}^{3}$, as well as the robot end-effector pose~${\mathbf{p} \in SE(3)}$, and the gripper position~$g \in \mathbb{R}$. The action space $\mathcal{A}$ parameterizes the Cartesian displacement~$\Delta \mathbf{x} \in \mathbb{R}^{3}$, the orientation displacement (axis–angle representation)~$\Delta \boldsymbol{\theta} \in \mathbb{R}^{3}$ of the end-effector, and the gripper position displacement~$\Delta g \in \mathbb{R}$.

\begin{figure*}[th!]
    \centering
    \includegraphics[width=0.69\linewidth]{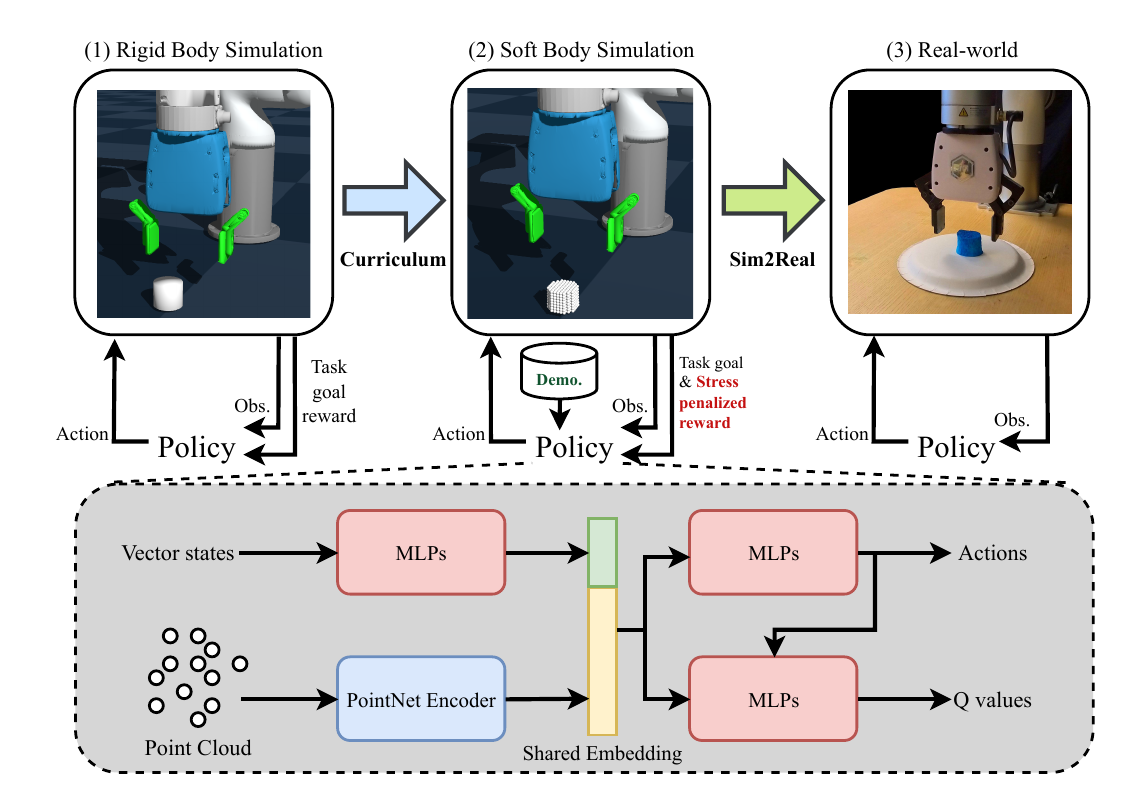}
    \caption{Overview of our stress-guided RL framework for deformable and fragile object manipulation. Training begins with (1) curriculum learning, where the agent first learns the task in rigid simulation before switching to soft-body simulation. (2) In the soft setting, stress-based guidance encourages safe manipulation, and expert demonstrations bootstrap learning. (3) Once converged, the policy is deployed zero-shot in the real world, with the setup closely matched to the simulation. The policy takes as input the segmented object point cloud and an 11-dimensional state vector: the 7-DoF end-effector pose pose~${\mathbf{p} \in SE(3)}$, the gripper width $g \in \mathbb{R}$ (in cm), and the centroid of the point cloud~$\mathbf{c}$.}
    \label{fig:method}
\end{figure*}

\subsection{Stress Metrics}\label{sec:method-prelim-stress-metrics}
We provide here a formal definition of stress.
We assume access to a Material Point Method (MPM) simulation of the deformable object, where the object is represented by $N$ particles.
For each particle $i$, we can compute a Cauchy stress tensor $\mathbf{C}^{(i)} \in \mathbb{R}^{3 \times 3}$. From this tensor, we derive the Von Mises stress~\cite{jong2009teaching} $\sigma^{(i)}$, a scalar quantity that aggregates stress information across all spatial directions.
More specifically, for the Moving Least Squares Material Point Method (MLS-MPM)~\cite{hu2018moving}, the Cauchy stress tensor $\mathbf{C}^{(i)}$ is computed during the Grid-to-Particle transfer. All stress values reported in this paper are in unit of Pa (N/$\text{m}^2$).

Now, to evaluate the level of stress experienced by an object using a single scalar, we propose to aggregate per-particle stress $\sigma^{(i)}$ in five ways: (i) mean stress, (ii) median stress, (iii) top percent mean stress, (iv) top percent median stress, and (v) maximum stress.
Suppose the particle stresses are ordered in a descending manner: $\sigma^{(1)} \geq \sigma^{(2)} \geq \sigma^{(i)} \geq \sigma^{(N)}$. 
The mean stress is given by $\bar \sigma = \Sigma_{i=1}^N\sigma^{(i)}/N$. The median stress $\hat \sigma$ is the median value over all particle stresses.
The top $K \in (0, 100)$ percent mean stress is determined as  ${\bar \sigma_{\text{top\{k\}}}=\Sigma_{i=1}^M\sigma^{(i)}/M}$, where $M=\lfloor KN/100 \rfloor$. Finally, we denote the maximum stress over particles as $\sigma_\text{max}$.


\section{METHOD}

We propose a stress-guided RL framework, as presented in Fig.~\ref{fig:method}. (1) Training begins with curriculum learning, where the agent first learns the task in rigid simulation. (2) After convergence in rigid setting, we switch to soft-body simulation and introduce stress-based guidance to promote safe manipulation, while expert demonstrations bootstrap learning. (3) Once the policy converges, it is deployed zero-shot in the real world, using a setup carefully matched to the simulation. The framework explicitly accounts for the object deformation by measuring the stress while performing the manipulation in a simulated environment. The learned policy can then be transferred to the real-world in a zero-shot manner, without any additional fine-tuning.

\subsection{Stress-Guided Reinforcement Learning}
We propose to use the stress obtained in simulation to guide policy learning, as illustrated in Fig.~\ref{fig:method}.
In this section, we start by describing the stress-penalized reward, followed by the details of policy training and architecture.


\subsubsection{Reward design}
The reward function consists of two terms: (i) Task success reward~$R_{\text{success}}$, 
and (ii) Stress-penalized reward~$R_{\text{stress}}$.
The task success reward encourages the robot to complete manipulation tasks such as picking and pushing.
$R_{\text{success}}$ also includes intermediate dense rewards to achieve each sub-goal of the task. 
For example, approaching the object as a precondition of a successful lifting is rewarded.
The stress-penalized reward combines the stress metrics introduced in Section~\ref{sec:method-prelim-stress-metrics}. This discourages actions that induce damage and promotes gentle behavior.

\begin{figure}
    \centering
    \includegraphics[width=0.945\linewidth]{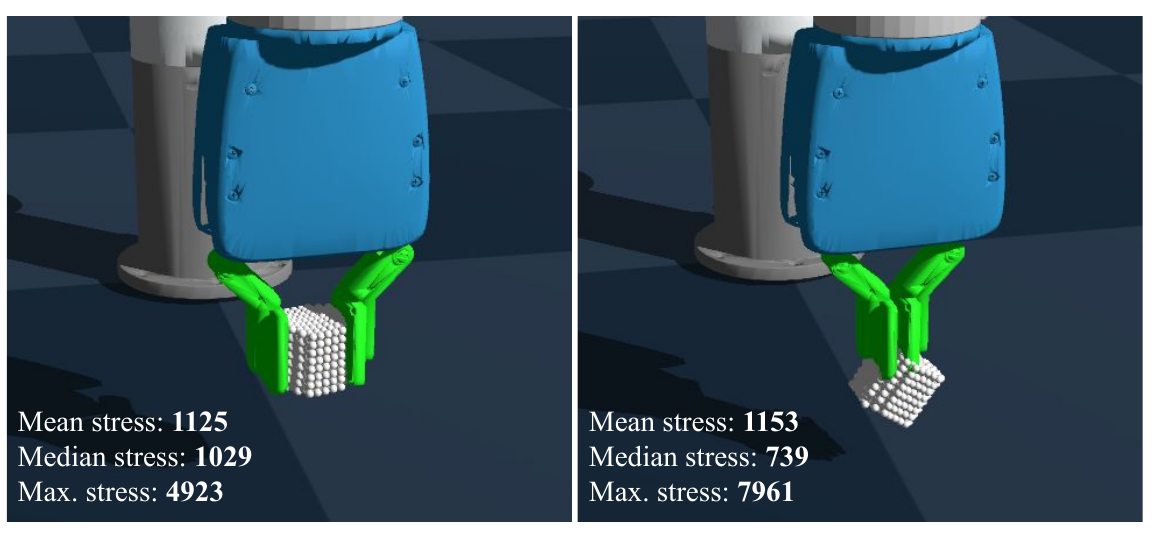}
    \caption{Global statistics (mean stress and median stress) may not reflect the local large deformation (as shown on the right), whereas max stress more reliably captures the information.}
    \vspace{-7pt}
    \label{fig:stress}
\end{figure}

The yield and fracture behavior of deformable and fragile objects is primarily dictated by the maximum stress applied across the object’s body. However, directly using the maximum stress as a penalty produces high-variance signals due to simulation randomness and policy execution variability. In contrast, global statistics such as mean or median stress can overly smooth the signal, masking localized high-stress regions. This is illustrated by an example shown in Fig.~\ref{fig:stress}: a firm grasp across the object (left), and a corner pinch (right). While both exhibit nearly identical mean and median stress values, the maximum stress reveals that the latter induces severe localized stress likely to cause damage.

To balance robustness and sensitivity, we propose a stress-penalized reward that captures localized peaks without excessive variance. The reward combines the mean stress~$\bar \sigma$ with the median of the top 10\% stress values~$ \hat \sigma_{\text{top10}}$ across particles, and applies a quadratic transformation to penalize high-stress regions more aggressively. The penalty is defined as

\begin{equation}
    \vspace{-3pt}
    R_{\text{stress}} = -\frac{1}{{\beta}}{(\alpha \hat \sigma_{\text{top10}} + (1-\alpha) \bar \sigma)^2},
\end{equation}

\noindent where $\alpha \in [0, 1]$ balances the penalty from $\hat \sigma_{\text{top10}}$ and $\bar \sigma$. The quadratic term amplifies the effect of large stresses, ensuring that occasional but dangerous peaks are penalized more severely. The scaling factor $\beta$ further sharpens the penalty once stress values exceed a threshold near $\beta$.


\subsubsection{Policy Training and Architecture} 
As shown at the bottom of Fig.~\ref{fig:method}, our policy encodes the object point cloud using PointNet~\cite{qi2017pointnet} and the robot state using an MLP. The two latent representations are concatenated to form a shared feature vector, which serves as input to both actor and critic networks, parameterized as MLPs. Policy optimization is performed using Soft Actor-Critic (SAC)~\cite{2018_haarnoja_sac}.

\subsection{Offline Demonstrations}
Incorporating a stress penalty into the reward can conflict with the primary task objective, since accomplishing the task requires exerting some stress. As a result, naively training policies with the stress penalty often degenerate into avoiding contact with the object to minimize stress, thereby failing to accomplish the task~\cite{huang2019learning} (Our experiment in Table \ref{tab:pickup-cylinder} validates this). To mitigate this issue, we leverage human demonstrations to bootstrap policy learning~\cite{luo2024serl,ball2023efficient}. 
We adopt the Reinforcement Learning with Prior Data (RLPD) framework with SAC as the backbone.
Following RLPD, we include offline demonstrations in the replay buffer alongside online rollouts using a symmetric sampling strategy. Specifically, we collect 20 demonstrations in simulation via keyboard tele-operation for each task, and each training batch consists of $50\%$ offline demonstrations and $50\%$ online experience. 
This integration of human prior knowledge improves sample efficiency, prevents convergence to suboptimal behaviors, and enables effective learning with the stress-penalized reward.

\subsection{Curriculum Learning}
Training with deformable objects is computationally demanding, as accurate simulation requires fine-grained timesteps. Compared to rigid-body dynamics, deformable simulations must employ substantially more substeps given the same control frequency to ensure stability, leading to significantly longer training times. To accelerate training while preserving task objectives, we make use of curriculum learning~\cite{narvekar2020curriculum}.
Here, we exploit the fact that (1) the task objective (pick-up/push) can be done on both rigid and soft object, (2) early learning stages do not require deformable object dynamics.

During the early stages of learning, the agent’s primary objective is to approach the object.
The behavior is largely independent of whether the object is rigid or deformable.
Motivated by this observation, we adopt the following curriculum learning strategy.
First, the policy is trained on a rigid object surrogate of the same shape.
Upon convergence, the object is switched to a deformable type (and thus soft body simulation is used). This approach 
improves overall training time efficiency while preserving task objectives.

\subsection{Mitigating the Sim-to-Real Gap}
To mitigate the impact of model mismatch between simulation and the real world of deformable and fragile objects, we employ domain randomization of key physical parameters.
In particular, the friction coefficient, Young’s modulus, and Poisson’s ratio of the object are randomized during training.
In addition, we randomize the initial pose of both the object and the robot end-effector, which helps to increase robustness against diverse initial configurations.

To account for perceptual uncertainty, we add zero-mean Gaussian noise to the observation, including the point cloud, its centroid, the end-effector pose, and the gripper position. The noise variance is set individually for each component.


\begin{figure}
    \centering
    \includegraphics[width=0.9\linewidth]{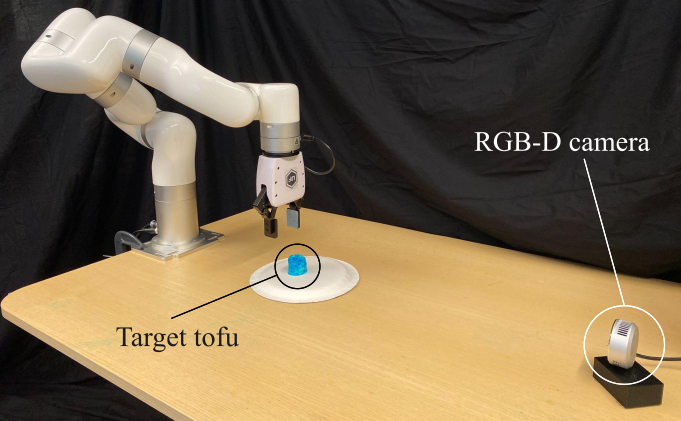}
    \caption{Experimental setup showing the robot manipulator, the target object, and the RGB-D camera used to provide observations to the control policy.}
    \label{fig:experiment-setup}
\end{figure}

\section{Experiments}
The experimental evaluation is structured around the following Research Questions (RQs):
(i) Does our stress-guided RL framework successfully learn to manipulate deformable and fragile objects, in terms of task success and avoiding damage to the object? (ii) How much does each of the design choices contribute to successful DFOM performance? (iii) Can the policies learned in simulation be successfully transferred to the real world in a zero-shot manner?

\subsection{Experimental setup}
We employ a UFactory xArm 7, a 7-DoF robotic manipulator with a standard parallel-jaw gripper (Fig.~\ref{fig:experiment-setup}). The deformable and fragile object used in our experiments is tofu, selected for its low stiffness and yield stress. This makes it a particularly challenging test case.
In simulation, we apply domain randomization over the tofu’s material properties, sampling Young’s modulus from $[5000, 10000]$ and Poisson’s ratio from $[0.325, 0.4]$, which encompasses the approximate physical characteristics of the real tofu~\cite{iopYoungsShould}.
In simulation, we use Genesis~\cite{zhou2024genesis} with the built-in Taichi soft body physical engine~\cite{hu2020difftaichidifferentiableprogrammingphysical}. We employ MPM rather than other methods, such as the Finite Element Method, as it offers more stable and accurate performance under large deformations typical in DFOM scenarios~\cite{de2020material}.
For real-world experiments, an external Intel RealSense L515 LiDAR camera is mounted in front of the robot to capture the object's point cloud. The tofu used in the experiments is shaped using 3D-printed cutters into cubes or cylinders and dyed with blue ink to facilitate segmentation from the background and robot via simple color thresholding.

\begin{figure*}[h]
    \centering
    \includegraphics[width=0.885\linewidth]{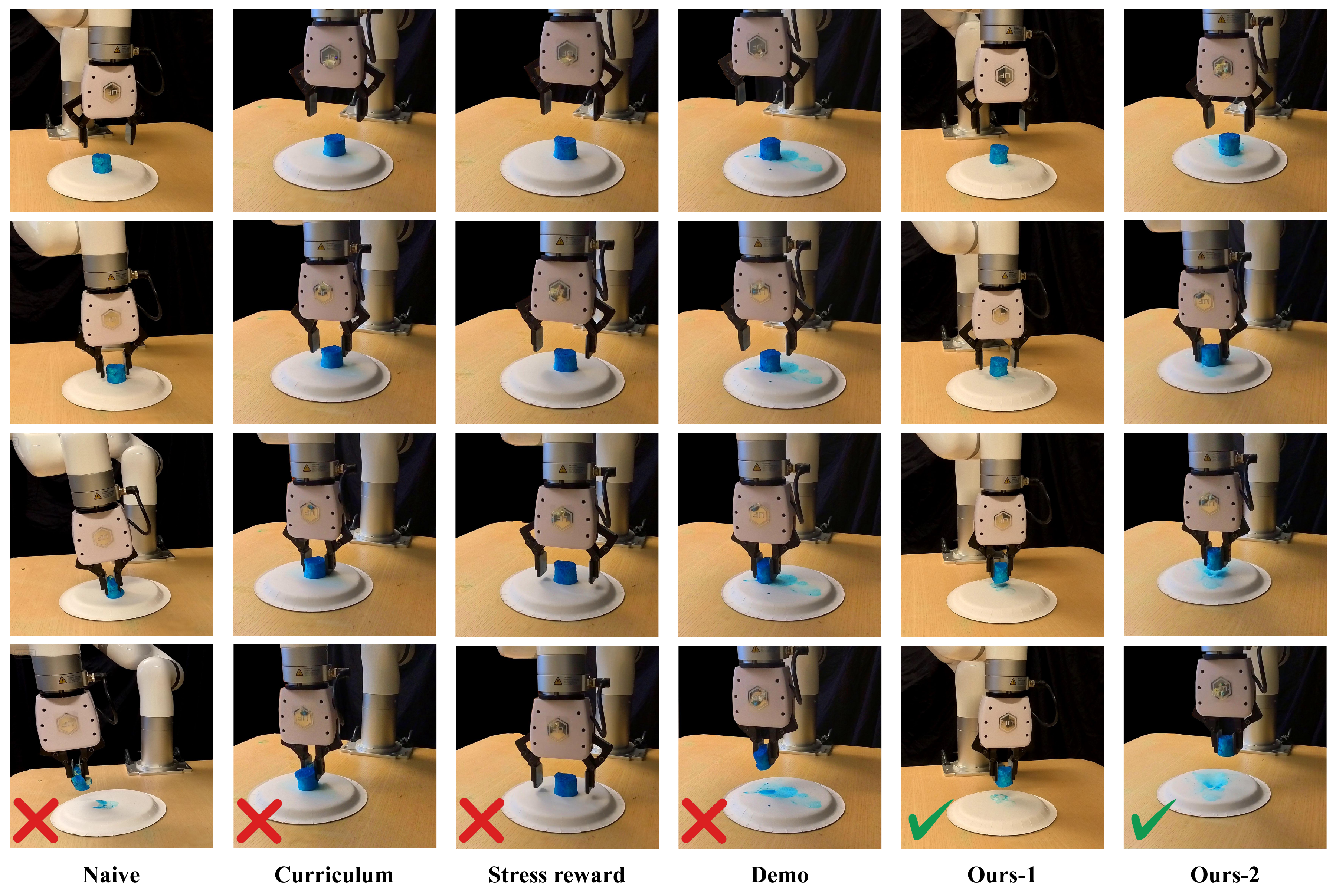}
    \caption{Qualitative results for the pick-up task of a cylindrical tofu. Each column illustrates a rollout with a different method. A green tick indicates task success without visible damage, while a red cross denotes either task failure or damage to the tofu.}
    \vspace{-5pt}
    \label{fig:pickup-cylinder}
\end{figure*}

\begin{figure*}[h]
    \centering
    \includegraphics[width=0.885\linewidth]{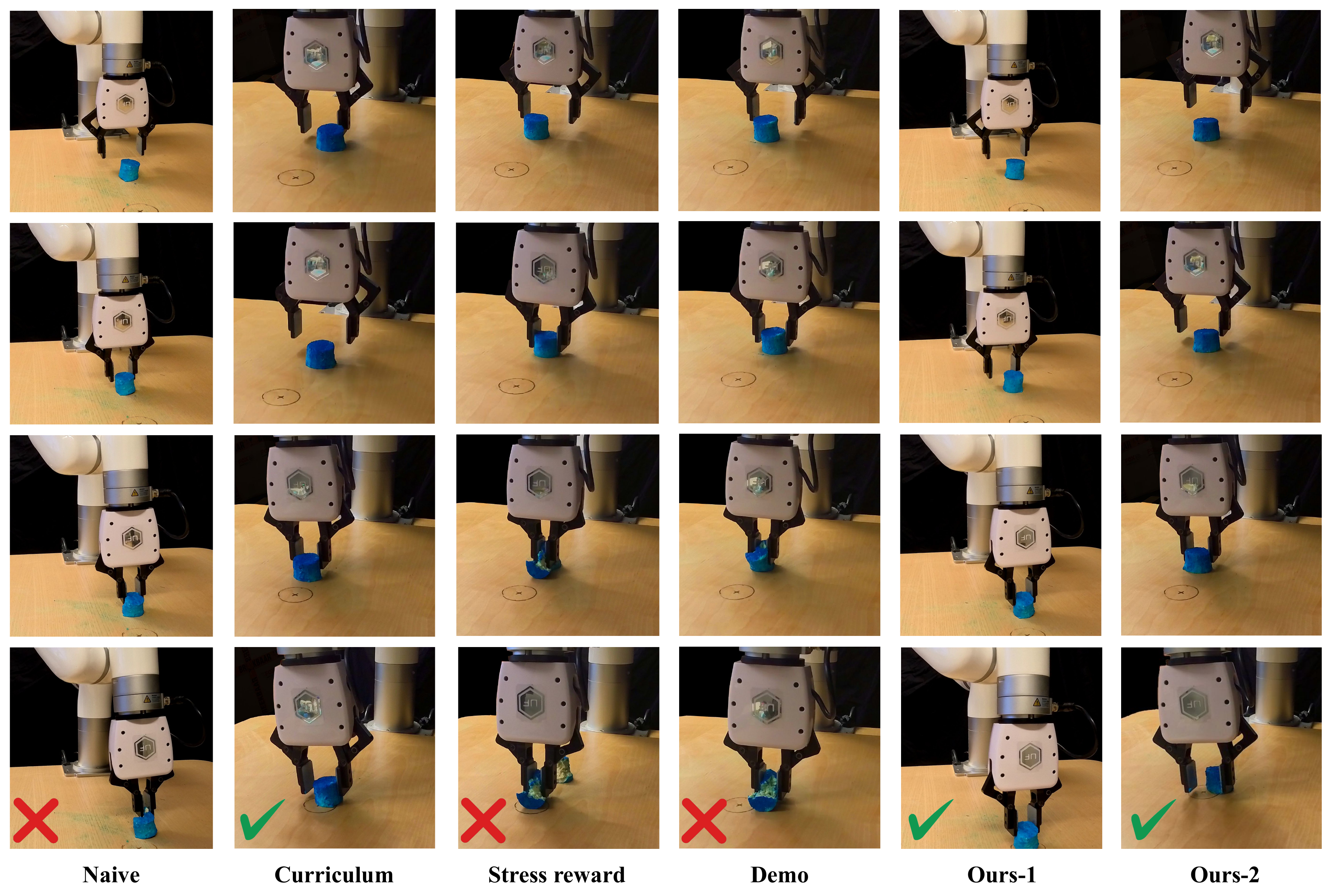}
    \caption{Qualitative results for the task of pushing a cylindrical tofu to a goal using different methods. The task begins with the gripper fully open, and the fingers may open or close during execution. Policies that damage the tofu often close the gripper tightly around the object while pushing it to the goal. In contrast, our methods close the gripper gently, enabling the tofu to be pushed reliably to the destination.
    }
    \label{fig:push-cylinder}
\end{figure*}

\begin{table}[!t]\centering
\caption{Evaluation on picking up a cylindrical tofu (simulation). \\
\textnormal{C: Curriculum, SPR: Stress-Penalized Reward, D: Offline Demonstrations, SR: Success Rate.}}
\label{tab:pickup-cylinder}
\rowcolors{2}{gray!10}{white}
\renewcommand{\arraystretch}{1.2}
\scriptsize
\begin{tabular}{c c c P{0.92cm} P{0.92cm} P{1.0cm} P{1.25cm}}\toprule
C & SPR & D & SR $\uparrow$  & $\bar{\sigma}$ $\downarrow$ & $\bar{\sigma}_{\text{top5}}$ $\downarrow$ & $\sigma_{\text{max}}$ $\downarrow$ \\\midrule
$\times$ & $\times$ & $\times$ & 0.90$\pm$0.04 & 2584$\pm$243 & 14974$\pm$1981 & 32879$\pm$5798 \\
\checkmark & $\times$ & $\times$ & 0.97$\pm$0.03 & 2527$\pm$139 & 12705$\pm$1001 & 25427$\pm$1508 \\
\checkmark & \checkmark & $\times$ & 0.90$\pm$0.12 & 2192$\pm$138 & 11891$\pm$1445 & 22575$\pm$4237 \\
$\times$ & \checkmark & $\times$ & 0.00 & N/A & N/A & N/A \\
$\times$ & \checkmark & \checkmark & 0.97$\pm$0.04 & 1966$\pm$150 & 9707$\pm$\textbf{319} & {17774$\pm$2245} \\
$\times$ & $\times$ & \checkmark & 0.95$\pm$0.03 & 2450$\pm$132 & 12934$\pm$1908 & 23888$\pm$2332 \\
\checkmark & $\times$ & \checkmark & \textbf{0.98$\pm$0.01} & 2343$\pm$\textbf{66} & 12122$\pm$380 & 24719$\pm$1190 \\
\checkmark & \checkmark & \checkmark & \textbf{0.98$\pm$0.01} & \textbf{1747}$\pm$67 & \textbf{9271}$\pm$459 & \textbf{16915$\pm$1048} \\
\toprule
\multicolumn{3}{c}{\texttt{BC}} &0.44$\pm$0.08 & 1910$\pm$167& 8066$\pm$301& 13413$\pm$602\\
\bottomrule
\end{tabular}
\renewcommand{\arraystretch}{1.0}
\end{table}

\begin{table}[!t]\centering
\caption{Evaluation on pushing a cylindrical tofu to a goal (simulation).}
\label{tab:push-cylinder}
\rowcolors{2}{gray!10}{white}
\renewcommand{\arraystretch}{1.2}
\scriptsize
\begin{tabular}{c c c c c c c}\toprule
C & SPR & D & SR $\uparrow$  & $\bar{\sigma}$ $\downarrow$ & $\bar{\sigma}_{\text{top5}}$ $\downarrow$ & $\sigma_{\text{max}}$ $\downarrow$ \\\midrule
$\times$ & $\times$ & $\times$ & 0.83$\pm$0.07 & 828$\pm$48 & 3859$\pm$186 & 7831$\pm$500 \\
\checkmark & $\times$ & $\times$ & \textbf{0.95$\pm$0.02} & 643$\pm$75 & 3011$\pm$144 & 5935$\pm$447 \\
\checkmark & \checkmark & $\times$ & 0.93$\pm$0.06 & 646$\pm$\textbf{19} & 3115$\pm$180 & 5972$\pm$377 \\
$\times$ & \checkmark & $\times$ & 0.72$\pm$0.09 & 854$\pm$92 & 3974$\pm$212 & 8322$\pm$310 \\
$\times$ & \checkmark & \checkmark & 0.83$\pm$0.04 & \textbf{546}$\pm$21 & \textbf{2443$\pm$99} & \textbf{4860$\pm$161} \\
$\times$ & $\times$ & \checkmark & 0.83$\pm$0.07 & 597$\pm$111 & 2797$\pm$480 & 5631$\pm$992 \\
\checkmark & $\times$ & \checkmark & 0.84$\pm$0.06 & 609$\pm$95 & 2913$\pm$269 & 6022$\pm$918 \\
\checkmark & \checkmark & \checkmark & 0.88$\pm$0.10 & 626$\pm$47 & 3088$\pm$274 & 6563$\pm$727 \\
\toprule
\multicolumn{3}{c}{\texttt{BC}} & 0.36$\pm$0.06 & 531$\pm$13 & 2343$\pm$42 & 4576$\pm$293 \\
\bottomrule
\end{tabular}
\renewcommand{\arraystretch}{1.0}
\vspace{-8pt}
\end{table}

\subsection{Tasks and Evaluation Metrics}
\label{sec:method-prelim-stress-metrics}
We evaluate two representative tasks: (i) \emph{Pick-up}: approach, gently grasp, and lift a piece of tofu. A pick-up is considered successful if the tofu is raised at least $9$~$cm$ above the table. (ii) \emph{Push}: move a piece of tofu on the table to a circular target area of radius $2$ $cm$, fixed at location $(x=47$~$cm, y=0, z=0)$ in the robot base frame. 
 
In simulation, we evaluate performance using both task success and stress-based metrics. The success rate is defined as the proportion of successful pick-ups or pushes, regardless of object damage. To quantify stress, we report the following statistics: the mean stress $\bar{\sigma}$, the mean of the top 5\% stress values $\bar{\sigma}_{\text{top5}}$, and the maximum stress over the object particles $\sigma_{\text{max}}$. For each of these stress metrics, we record the maximum value observed over the entire episode as the final quantity. In the case of task failure, stress values are excluded from evaluation to avoid biasing the averages.


\subsection{Implementation Details}

For pick-up tasks, we combine approach $r_a$, lift $r_l$, distance-to-goal $r_g$, and success $r_s$ rewards. We define $r_a = \exp(-20d)$, where $d$ is the Euclidean distance from the object to the tool center position; $r_l = \text{clip}(h/0.9, 0, 1)$, where $h$ is the object height; $r_g = \exp(-20d_\text{goal})$, where $d_\text{goal}$ is the Euclidean distance from the object to the goal location; The success reward $r_s$ is binary, indicating whether the object reaches the goal height, with $2$~$cm$ threshold. Reward scales are $0.3$, $0.7$, $1.0$, and $2.0$ for $r_a$, $r_l$, $r_g$, and $r_s$ respectively. Push tasks use the same formulation, except for (1) no $r_l$, (2) $r_s$ and $r_g$ depend on planar distance to the goal. Stress penalties use $\beta = 6000$, $\alpha = 0.8$, and a scale of $5 \times 10^{-5}$.  

To ensure reproducibility and support future research on DFOM, we will release our code, trained checkpoints, demonstrations, hyper-parameters, and links to the tofu \& foam used in real-world experiments on our 
\href{https://sites.google.com/view/gentle-manipulation}{project website}. 

\begin{table}[!t]\centering
\caption{Evaluation on picking up a cubic tofu (simulation).}
\label{tab:pickup-cube}
\rowcolors{2}{gray!10}{white}
\renewcommand{\arraystretch}{1.2}
\scriptsize
\begin{tabular}{c c c c c c c}\toprule
C & SPR & D & SR $\uparrow$  & $\bar{\sigma}$ $\downarrow$ & $\bar{\sigma}_{\text{top5}}$ $\downarrow$ & $\sigma_{\text{max}}$ $\downarrow$ \\\midrule
$\times$ & $\times$ & $\times$ & \textbf{0.98$\pm$0.03} & 2354$\pm$203 & 11505$\pm$747 & 20018$\pm$2575 \\
$\times$ & \checkmark & \checkmark & 0.97$\pm$\textbf{0.03} & \textbf{1505$\pm$61} & \textbf{8448}$\pm$480 & \textbf{12766}$\pm$706 \\
\checkmark & \checkmark & \checkmark & \textbf{0.98$\pm$0.03} & 1981$\pm$155 & 9376$\pm$\textbf{314} & 15107$\pm$\textbf{646} \\
\bottomrule
\end{tabular}
\renewcommand{\arraystretch}{1.0}
\end{table}

\begin{table}[!t]\centering
\caption{Evaluation on pushing a cubic tofu (simulation).}
\label{tab:push-cube}
\rowcolors{2}{gray!10}{white}
\renewcommand{\arraystretch}{1.2}
\scriptsize
\begin{tabular}{c c c c c c c}\toprule
C & SPR & D & SR $\uparrow$  & $\bar{\sigma}$ $\downarrow$ & $\bar{\sigma}_{\text{top5}}$ $\downarrow$ & $\sigma_{\text{max}}$ $\downarrow$ \\\midrule
$\times$ & $\times$ & $\times$ & 0.76$\pm$0.06 & 719$\pm$104 & 3154$\pm$267 & 4848$\pm$307 \\
$\times$ & \checkmark & \checkmark & 0.68$\pm$\textbf{0.01} & \textbf{524}$\pm$\textbf{61} & \textbf{2502}$\pm$195 & \textbf{3720}$\pm$329 \\
\checkmark & \checkmark & \checkmark & \textbf{0.84$\pm$0.01} & 696$\pm$\textbf{61} & 3102$\pm$\textbf{98} & 4612$\pm$\textbf{246} \\
\bottomrule
\end{tabular}
\renewcommand{\arraystretch}{1.0}
\end{table}

\begin{table}[!t]\centering
\caption{Evaluation on picking up a cubic foam (real world).}\label{tab:foam}
\rowcolors{2}{gray!10}{white}
\renewcommand{\arraystretch}{1.2} 
\scriptsize
\begin{tabular}{c c c c}\toprule
&Pick-up Success Rate $\uparrow$ &Water Loss \%  $\downarrow$\\\midrule
\texttt{BC}  &0.40$\pm$0.51 & \textbf{9.52}$\pm$3.88 \\
\texttt{Naive} &\textbf{1.00}$\pm$\textbf{0.00} &41.90$\pm$5.40 \\
\texttt{Ours-1} &\textbf{1.00}$\pm$\textbf{0.00} &10.00$\pm$\textbf{1.50} \\
\texttt{Ours-2} &0.90$\pm$0.31 &21.69$\pm$7.12 \\
\bottomrule
\end{tabular}
\vspace{-18pt}
\end{table}

\subsection{Evaluation in Simulation}
We evaluate whether the proposed method manipulates fragile objects effectively while minimizing the stress applied stress.
In addition, we quantify the contribution of three design components: (i) curriculum learning, (ii) stress-penalized reward, and (iii) demonstrations. We consider the \textit{Pick-up} and \textit{Push} tasks on tofu objects with two distinct geometries: cylindrical and cubic. We consider two variants of our method reported as \texttt{Ours-1} (stress-penalized reward \& demonstrations) and \texttt{Ours-2} (full method: curriculum \& stress-penalized reward \& demonstrations). We further integrate in the evaluation the following baselines and ablations: \texttt{Naive} (vanilla RL without curriculum, stress-penalized reward, or demonstrations), \texttt{Curriculum} (only curriculum), \texttt{Stress reward} (only stress-penalized reward), \texttt{Demo} (only demonstrations), and Behavior Cloning \texttt{BC}~\cite{torabi2018behavioral}. For all methods, including \texttt{BC}, we use the same 20 demonstrations per task and object, and identical reward hyper-parameters.

Tables~\ref{tab:pickup-cylinder} \& \ref{tab:push-cylinder} present the full ablation on cylindrical tofu (primary benchmark). Tables~\ref{tab:pickup-cube} \& \ref{tab:push-cube} report a reduced evaluation on cubic tofu to probe shape generalization.
The results are reported over 3 random seeds.


The results show a trend of decreasing applied stress for our proposed method while retaining a high success rate, whereas baselines that do not consider stress consistently induce the highest stress on the object. More specifically, the key findings are:
(i) The \texttt{Naive} method can achieve moderate task success but typically induces high stress, resulting in frequent tofu damage.
(ii) The \texttt{Stress reward} makes the policy overly conservative, preventing contact with the tofu and leading to poor task success.
(iii) In both pick-up and push tasks (see Table~\ref{tab:pickup-cylinder} and \ref{tab:push-cylinder}), relying solely on demonstrations (\texttt{Demo}) is insufficient to achieve a low-stress manipulation policy, even though the demonstrations are carefully collected to minimize stress.
In contrast, \texttt{Ours-1} — which adds stress-penalized rewards in addition to demonstration — yields significantly better results.
This underscores the necessity of incorporating stress-based rewards, even when high-quality demonstrations are available.
(iv) Curriculum and demonstrations both serve as effective bootstrapping mechanisms, with demonstrations additionally providing human knowledge for gentle handling.
Interestingly, \texttt{Ours-1} often matches or even surpasses \texttt{Ours-2} (all three) in minimizing stress.
(v) Compared to our proposed approaches, while \texttt{BC} reduces stress, its low task success rate makes it less effective considering the binary objectives.

Altogether, these results address RQs (i) and (ii), confirming that our stress-guided RL framework enables effective manipulation of fragile objects under reduced stress, with the stress-penalized reward as a necessary component, and curriculum learning demonstrations providing complementary improvements.

\begin{figure}[t!]
    \centering
    \includegraphics[width=0.915\linewidth]{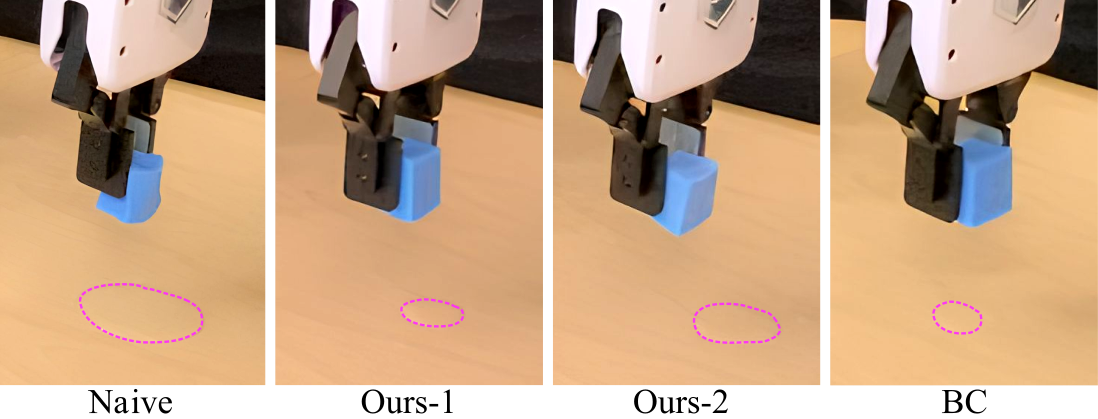}
    \caption{Qualitative results for the water loss from a fully water-absorbed foam after executing the pick-up task with different methods. The dropped water is highlighted in purple, a smaller marked area indicates better performance. While \texttt{BC} exhibits the least water loss, its pick-up success rate is less than half that of \texttt{Ours-1} and \texttt{Ours-2}.}
    \vspace{-12pt}
    \label{fig:foam-quality}
\end{figure}

\subsection{Real-World Experiments}

We assess the zero-shot transfer capability of our method to real-world DFOM, evaluating both quantitatively and qualitatively the stress applied to the objects despite not having direct access to it.\subsubsection{Quantitative evaluation}
We assess the gentleness of the grasp using an absorptive foam block saturated with water. The foam is weighed before and after each pick-up trial, and the difference (i.e., water loss due to squeezing) is normalized by the initial absorbed water.
Each trial starts with the same initial water content (21 g, excluding the foam’s own weight). Policies are evaluated using four methods across 10 trials 
each: \texttt{Naive}, \texttt{Ours-1}, \texttt{Ours-2}, and \texttt{BC}. The results are summarized in Table~\ref{tab:foam}, with a qualitative visualization shown in Fig.~\ref{fig:foam-quality}. 

{Key takeaways}: (i) Both \texttt{Ours-1} and \texttt{Ours-2} achieve high pick-up success rates with substantially reduced water loss, indicating gentler manipulation compared to \texttt{Naive}. \texttt{BC} also minimizes water loss, but achieves only $40\%$ success. This suggests that more than 20 demonstrations may be needed, which requires a higher data collection cost than our bootstrapped RL method. (ii) Policies trained on tofu transfer to foam, due to the similarity in their physical properties.

\subsubsection{Qualitative evaluation}
We qualitatively evaluate \texttt{Naive}, \texttt{Curriculum}, \texttt{Stress reward}, \texttt{Demo}, \texttt{Ours-1}, and \texttt{Ours-2} across the pick-up \& push tasks. Results for cylindrical tofu are shown in Fig.~\ref{fig:pickup-cylinder} \& Fig.~\ref{fig:push-cylinder}. Our methods (\texttt{Ours-1} \& \texttt{Ours-2}) consistently achieve the task objectives while preserving the integrity of the tofu. In contrast, other methods either complete the task at the cost of damaging the tofu or fail to complete it altogether. These findings are consistent with the simulation results, demonstrating successful zero-shot sim-to-real transfer. 
In the spirit of eco-friendliness and given the clear performance difference of the methods, we avoid repeating physical experiments with tofu to reduce food waste.


\section{Conclusions}
We proposed a stress-guided RL method for deformable and fragile object manipulation.
Our approach enables learning a damage-aware policy using only visual input, which can be transferred to the real world in a zero-shot manner. 
To achieve this, we introduced a stress-penalized reward in simulation and bootstrapped policy learning through curriculum and offline demonstrations.
We evaluated our method in both simulated and real-world tasks, including the pick-up \& pushing of fragile objects.
Our results showed that, by including the proposed design components, our approach was able to successfully perform these tasks while minimizing damage to the objects. 
This work highlights the importance of integrating physical priors, such as the stress when manipulating fragile objects, in guiding learning approaches to enable more reliable and gentle manipulation. 

While our method mitigates the sim-to-real gap through domain randomization, discrepancies between simulated and real-world soft body modeling and dynamics remain a challenge.
Future work could explore training RL policies in the real-world to further reduce this gap. 
Additionally, the proposed approach could be extended to handle generalization across a broader range of objects, including food items and other soft materials. 
Finally, we identified the need for a standardized benchmark in this area, which would facilitate more consistent evaluations of future methods.

\bibliographystyle{IEEEtran}
\bibliography{IEEEabrv, ref}





\end{document}